# UN MODELE OBJET POUR LA REPRESENTATION
# DE CONNAISSANCES EMPIRIQUES

Joël COLLOC    Danielle BOULANGER
Laboratoire Informatique
IAE Université Lyon 3 – 1, rue de l'Université –69007 Lyon – France.

## Résumé

Nous proposons un nouveau modèle orienté objet exprimant l'aspect statique et dynamique des connaissances appartenant à des domaines variés.
Ce modèle offre un double niveau conceptuel
- Le niveau Interne, le plus important, établit * la structure d'un objet en terme de hiérarchie de sous-objets le composant,
. l'évolution de cette structure à l'aide de fonctions dynamiques,
. la comparaison d'objets de même type à l'aide de fonctions d'évaluation internes.
il définit un héritage interne ascendant, des propriétés des sous-objets parties
vers l'objet unique.
- Le niveau Externe représente :
. l'environnement d'un objet (les autres objets),
. il définit la notion de type d'objet et de hiérarchie de types et ce partant d'un héritage externe simple du type vers les sous-types.

## Abstract

We are currently designing an object oriented model which describes static and dynamical knowledge in différent domains.
It provides a twin conceptuel level :
- the internal level proposes
. the object structure composed of sub-objects hierarchy,
. structure évolution with dynamical fonctions,
. same type objects comparaison with évaluation fonctions.
It uses multiple upward inheritance from sub-objects properties to the Object.
- the external level describes
. object environment,
. it enforces object types and uses external simple inheritance from the type to the sub-types.

## 1. Introduction : pourquoi un modèle orienté objet ?

Nous présentons un modèle orienté objet destiné à la construction de bases de connaissances, exploitables par des systèmes experts raisonnant en univers incertain: sciences humaines, biologie, médecine, géologie... Il décrit l'aspect statique et dynamique des informations.
- L'imprécision, l'incertitude des données expérimentales ont conduit à une modélisation à l'aide de fonctions continues.
- Le modèle autorise la prise en compte au même moment de tous les objets concernant la résolution d'un problème donné.
Son but est d'exprimer la connaissance sous forme d'objets exploitables par des systèmes experts examinant les hypothèses non pas successivement mais simultanément au cours du même cycle de raisonnement.
Ce modèle privilégie un raisonnement de type associatif au détriment d'un raisonnement canalaire.

### 1.1. Evaluation des modèles à règles de production

Les règles de production sont bien systématisées et de nombreux systèmes les utilisent. Par contre, elles sont incapables d'exprimer des valeurs continues même si des coefficients peuvent donner cette impression.
Les systèmes experts font des choix binaires en déclenchant ou non des règles, le raisonnement est donc de nature canalaire puisqu'à un moment donné, un seul fragment de connaissance est confronté avec le problème à résoudre.
Gérer l'ordre de déclenchement des règles nécessite l'implantation de méta-règles insuffisantes à empêcher la survenue de boucles de raisonnement liées à la transitivité, ou à maîtriser des règles contradictoires provenant d'experts différents. (Flory et Bertin 1986). Enfin, il est difficile d'exprimer l'aspect dynamique de la connaissance à l'aide de règles et notamment qu'un fait puisse évoluer, au cours de la consultation du Système Expert, et même disparaître.

## 2. Présentation du modèle

### 2.1. Objectifs

Le modèle à objets décrit dans ce travail est destiné à représenter de la connaissance incertaine, approximative, variable dans le temps et basée sur l'expérience des hommes. Il favorise la réalisation de tâches d'expertise ou de simulation tout en permettant l'accès à la base de données objets.
Ce modèle se distingue par les points suivants :
1 - Il établit un double niveau conceptuel : *Niveau Conceptuel Interne*
et *Niveau Conceptuel Externe.*
2 - Il décrit non seulement l'aspect statique : les faits, mais également leur évolution spontanée dans le temps.
Nous représentons les caractéristiques dynamiques à l'aide de fonctions continues ou discrètes, internes aux objets, exprimant d'une part une modification des liens de structure et d'autre part un changement de la valeur de référence de leurs attributs.

3 - La généralisation bâtit une hiérarchie de types, située au niveau externe. Elle autorise la transmission des caractéristiques du type aux objets de ses sous-types, nous l'appelons héritage externe. Cet héritage parcourt la hiérarchie de haut en bas : des types vers leurs sous-types. L'utilisation de l'héritage multiple doit être très prudente. (Boulanger et Morejon 1988).

4 - Nous différencions type et classe, la classe étant l'ensemble des instances et le type d'objet expriment leurs caractéristiques.

5 - L'agrégation est privilégiée dans notre modèle elle intervient au niveau interne des objets qui sont composés de sous-objets. Nous appelons *Composition d'objets* l'utilisation récursive de l'agrégation pour former un objet à l'aide de ses sous-objets.
La Composition d'objets bâtit une *hiérarchie interne* sous forme d'une partition et définit *l'héritage interne.*

6 - Exploitation de l'héritage interne.
L'objet unique hérite automatiquement de toutes les caractéristiques statiques et dynamiques de ses sous-objets qui disposent de leur propre évolution.
L'objet unique hérite non seulement des attributs et des fonctions des sous-objets mais également de leurs valeurs.

7 - Nous exprimons la connaissance déductive grâce à des *fonctions d'évalua-tion* de deux types :
  . Les fonctions d'évaluation *propre à l'objet,* continues ou discrètes situées au niveau interne, dans l'objet unique. Elles solutionnent le problème de l'incertitude des seuils par le calcul d'une distance (Colloc et Boulanger 1987).
  . Les fonctions d'évaluation *globales*, le plus souvent discrètes, intervenant au niveau externe et liées à un type d'objet qu'elles contribuent à définir.

8 - Les *attributs* résident au niveau interne, ce ne sont pas des objets.

9 - La perception des objets : chaque objet contient des sous-objets et est lui même contenu dans un autre objet de taille plus importante.

Pour chaque application, il faut choisir une taille d'objet appropriée. C'est ce que nous appelons *l'effet zoom* (figure 1).

## 2.2. Le niveau internet et le niveau externe

### 2.2.1. L'opposition du niveau interne au niveau externe

Un observateur considère soit la structure INTERNE, soit l'environnement EXTERNE d'un objet.

La structure INTERNE représente le contenu d'un objet et de ses sousobjets composants, formant une partition.

L'environnement EXTERNE traduit les rapports d'un objet avec les autres.

Les classifications sont EXTERNES à l'objet et existent indépendamment de lui. Elles peuvent d'ailleurs être multiples et même se chevaucher puisqu'un même objet peut être classifié de plusieurs manières différentes. L'arborescence d'une classification est indépendante de celle représentant un objet physique au niveau interne.

### 2.2.2. La définition d'un objet de référence ou objet unique,

Nous percevons les objets comme emboîtés les uns dans les autres.

Il faut établir, selon les applications, une unité, correspondant à un type d'objet de référence, nous appelons chaque instance de ce type *objet unique*.

Tous les objets de taille supérieure ou égale appartiennent au milieu externe, tandis que les plus petits sont internes à l'objet unique et contribuent à sa formation, ce sont les sous-objets.

*Exemple* : En biologie, si l'unité choisie est une cellule hépatique, ses organites (ex: noyau) sont internes et par conséquent sont des sous-objets de l'objet de référence cellule. Les autres cellules, organes, appareils sont externes.

Les instances du type d'objet de référence résident au niveau 0.

Un objet est tout d'abord composé de connaissances structurales et factuelles définissant son aspect statique.

Des procédures et des fonctions du temps expriment son évolution, sa dynamique tandis que d'autres fonctions ou méthodes décrivent les actions qu'il peut exécuter.

Dans le modèle, un objet délimite un milieu interne avec ses lois, protégé de l'environnement appelé milieu externe.

L'encapsulation garantit cette protection, elle interdit l'accès au contenu de l'objet en dehors de l'utilisation de messages. (Brodie 1981), (Laurent 1987), (Dittrich et Lorie 1985), (Methfessel 1987), (Zaniolo Ait-kaci et Beech 1986), (Barbedette et Richard 1987).

### 2.2.3. Une perception adaptée des objets : l'effet Zoom

A un moment donné, le système considère un seul type d'objet de référence ou niveau 0, toutefois, il peut traiter successivement des objets de taille différente. Nous appelons cette adaptabilité du niveau de travail effet Zoom.

Tout sous-objet peut être vu, à un moment donné, au niveau 0, ses propres sous-objets constituant son milieu interne. Nous définissons la notion de gros-sissement permettant à un expert de définir la taille des objets utile à la réso- lution d'un problème.

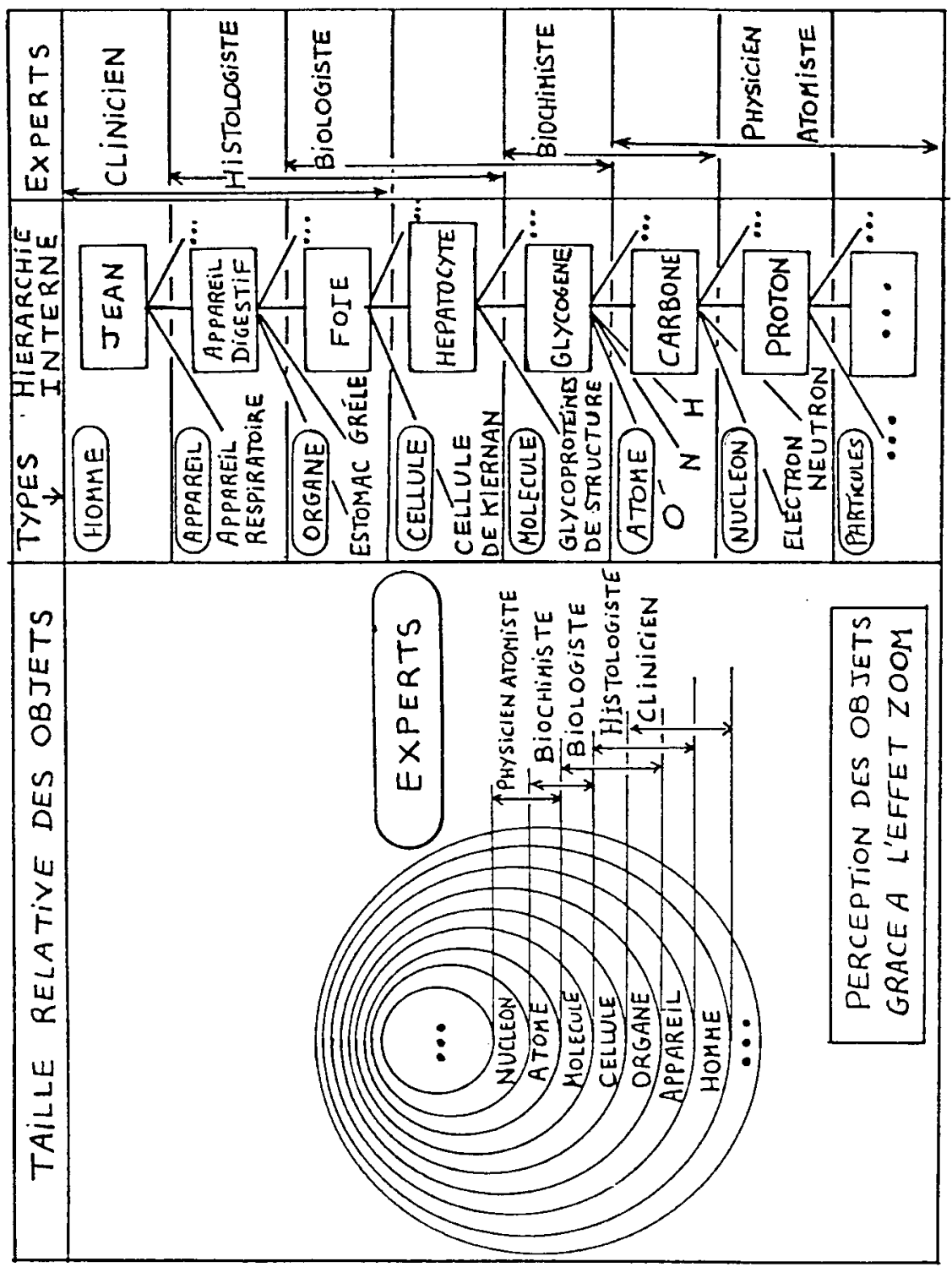

Figure 1 : Perception des objets grâce à l'effet zoom

## 2.2.4. Les sous-objets forment une arborescence interne.

Selon la précision désirée, l'expert représente les objets du type de référence, à l'aide d'un certain nombre de niveaux imbriqués de sousobjets. La description du niveau interne d'un objet est récursive. Cette arborescence interne est propre à chaque instance du type d'objet de référence.

L'objet de niveau 0 est une instance dont la structure et les contraintes sont conformes au type d'objet de référence choisi par l'expert, c'est un objet unique résidant au sommet de l'arborescence interne des sous-objets dont il est en fait constitué.

Au niveau externe, l'objet unique appartient à la classe associée au type d'objet de référence. Chaque sous-objet est nécessairement d'un type différent du type de l'objet unique et appartient donc à une autre classe.

## 2.2.5. Composition d'objets appartenant à des types distincts.

La Composition d'objets construit un objet complexe unique en appliquant récursivement l'agrégation sur des sous-objets. Chacun d'eux résulte de l'agrégation de sous-objets plus simples.

Soit TYa et TYb deux types d'objets, AQa et AQb l'ensemble de leurs attri- buts qualifiants respectifs (voir § Les attributs qualifiants : leur rôle interne),
Cla et Clb les classes d'objets associées.

On définit R une relation de composition sur l'ensemble des types, telle que TYc $\in$ TY : TYc = TYa R TYb

On montre sur un exemple que : CLc diffère de CLa $\cap$ CLb et AQc = AQa $\cup$ AQb. De plus on remarque que CLc n'est ni inclu dans CLa, ni dans CLb.

Par conséquent, le type résultant TYc n'est sous-type d'aucun des deux types impliqués dans la composition.

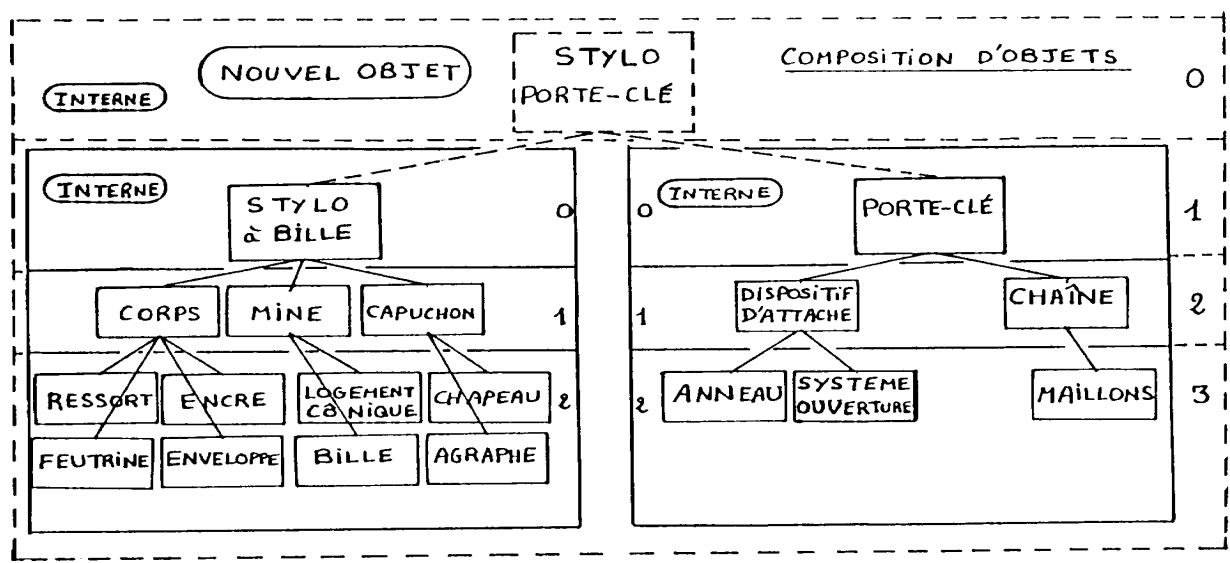

Figure 2 : exemple de composition d'objets

TYc comporte un niveau supplémentaire par rapport aux types TYa et TYb :
le nombre d'attributs qualifiants a augmenté.

Un stylo-gomme n'est plus tout à fait un stylo ni une gomme, il a des pro-priétés supplémentaires provenant de l'union des ensembles des attributs quali-fiants, lui interdisant de faire partie des classes CLa et CLb.

*Au niveau* INTERNE,les objets de la classe CLb et de la classe CLa sont sous-objets des objets de la classe CLc.

Un nouveau type d'objet est créé en combinant les types TYa et TYb. Le nombre de niveaux des objets des types impliqués n'est pas toujours identique comme dans l'exemple.

Les attributs qualifiants sont hérités de bas en haut, conformément au type d'objets associé. En composant successivement des types d'objets, on en crée de nouveaux de plus en plus complexes.

Le nombre d'attributs qualifiants, de niveaux, et par conséquent de con-traintes croît proportionnellement.

On définit la relation « EST - COMPOSEE – DE » notée o sur *l'ensemble des objets*. Pour tout OBJc $\in$ CLc, il existe au moins un OBJa $\in$ CLa et il existe au moins un OBJb $\in$ CLb / OBJc = OBJa o OBJb.

La relation o réalise la Composition de deux sous-objets en un objet unique, elle intervient au *niveau interne* du modèle.

En conclusion : la Composition d'objets de deux types entraîne la forma-tion d'un troisième objet qui n'est instance d'aucun des deux types d'origine et dont l'ensemble des attributs qualifiants est l'union des ensembles des attributs qualifiants des types premiers.

Le type de ce nouvel objet résulte de la composition des types respectifs des objets impliqués dans la composition d'objet.

Le nouveau type décrit des objets plus complexes car le cardinal de l'ensemble des attributs qualifiants a augmenté.

Au niveau Interne : les objets du nouveau type ont comme sous-objets des objets appartenant aux deux types d'origine impliqués dans la Composition.

## 2.3. Le niveau Interne

### 2.3.1. Le rôle de la composition d'objets

Certains auteurs (Barbedette et Richard 1987), (Banerjee, Chou et Garza 1987), (Hornick et Zdonik 1987) définissent dans leur modèle la notion d' « ob-jet composite ». Bien que proche de la notion d'objet unique, nous relevons un certain nombre de différences :

1 - La notion de niveau interne et externe n'existe pas, de sorte que les objets composites" ne représentent qu'une forme particulière de la hiérarchie de classe qui est privilégiée. Alors que dans le modèle présenté, le niveau interne est le plus important.

Le niveau externe ne sert qu'à implanter de nouveaux types, à réaliser la communication entre des objets de nature différente et l'interface avec l'utili-sateur.

2 - Nous utilisons l'association 'EST_PARTIE_DE' au niveau interne pour agréger les sous-objets d'une des instances d'un type de référence et non pas au niveau externe sur une classe d'objets, car il n'existe généralement pas de partition au niveau de la hiérarchie de classe. (voir § Généralisation et Hiérarchie de types et de classes : hiérarchie externe).

3 - La hiérarchie de composition située au niveau interne est instanciée de *bas en haut*, c'est à dire des sous-objets vers l'objet unique et l'héritage des attributs a lieu *dans le même sens*, ce qui représente une des particularités du modèle.

La modification d'un des sous-objets retentit donc immédiatement sur l'objet unique dont il fait partie.

4 - La partition formée par les sous-objets à chaque niveau, nous permet d'implanter un héritage interne multiple sans les conflits rencontrés en utilisant l'héritage externe multiple.

5 - Le modèle offre la possibilité d'agir d'une manière générique sur toutes les instances en s'adressant à un type d'objet mais également de manière spécifique sur un des sous-objets composants sans interférer sur les autres.

## 2.3.2. La notion de sous-objet

Un sous-objet est un objet appartenant au niveau interne, nécessairement inclus dans la structure de l'objet unique. A ce titre il est relié à lui directement ou indirectement (par l'intermédiaire d'autres sous-objets), par l'association « EST_PARTIE_DE ».

Sa propre structure est soit simple (uniquement constituée d'attributs) soit complexe composée d'autres sous-objets.

Dans le modèle, le niveau des sous-objets est >= 0. (un objet est sous-objet de lui-même).

Un sous-objet peut lui même avoir un ou plusieurs sous-objets.

> De manière générale : Si nl est le niveau d'un objet et si n2 est le niveau d'un sous-objet de cet objet alors on a ni <= n2.

Les sous-objets composant l'objet peuvent comporter des attributs, des fonctions d'évaluation permettant de comparer cette instance avec une autre, des fonctions de structure chargées d'exprimer l'incertitude ou l'inconstance de l'implication d'un sous-objet dans la formation de l'objet.

## 2.3.3. Structure d'un objet unique : la hiérarchie interne

Un objet unique est représenté par une hiérarchie de composition interne établissant sa structure ainsi que celle de tous les sous-objets impliqués dans sa constitution.

On peut construire des objets simples (ne comprenant que des attributs et des fonctions) puis en composant ces derniers, créer des objets uniques de types de plus en plus complexes (voir § Composition d'objets appartenant à des types distincts).

On peut aussi décomposer un objet existant en ses constituants, l'expert devra d'abord établir la taille de l'objet unique considéré et la précision (en nombre de niveaux) avec laquelle il désire décrire l'objet.

Il faut ensuite recenser les sous-objets et établir la hiérarchie de composition. Enfin, il faut créer les sous-objets les plus simples (feuilles de l'arborescence) puis les agréger en sous-objets de plus en plus complexes, jusqu'à réaliser l'objet unique fixé comme but.

Selon la structure définie par le type d'objet, un objet de niveau 0 est consti-tué de sous-objets de niveau niv où niv varie de 1 à Nbnv. (Le maximum Nbnv est spécifié par l'expert : concepteur du type d'objet. Un objet du niveau niv est formé de sous-objets de niveau niv + l.

Les sous-objets de niveau niv + l sont unis à l'objet de niveau niv par des liens traduisant le fait qu'ils concourent à le former. Ces liens représentent l'association "EST-PARTIE-DE".

La prépondérance de certains liens sur d'autres s'exprime grâce à une fonc-tion interne, le plus souvent discrète mais parfois continue. (Voir § Evolution de la structure interne des objets).

Dans les domaines, comme en médecine, où la connaissance est incertaine et figurée par des objets abstraits, l'expert attribue à chaque lien un coefficient de fiabilité traduisant le degré de certitude qu'un sous-objet est une partie de l'objet. L'expert définit pour chaque objet des attributs, des fonctions internes destinées à évaluer son état.

Enfin, l'objet comprend des explications, questions, graphismes, expliquant son rôle à l'utilisateur, facilitant le dialogue et la saisie de valeurs par des péri-phériques variés (clavier, souris, table à digitaliser... ).

La hiérarchie interne d'agrégation réalise une partition à chaque niveau de l'arborescence, car un objet unique est entièrement formé de ses sous-objets et chacun d'entre eux possède cette même qualité. Cette faculté permet d'implanter *un héritage interne multiple et surtout sans conflit.*

### 2.3.4. Un héritage interne multiple et ascendant.

Au niveau INTERNE, du fait de la partition, le sens de transfert des attributs a lieu de BAS en HAUT, des sous-objets vers l'objet unique référencé au niveau 0.

Chaque attribut d'un sous-objet devient immédiatement par HERITAGE ASCENDANT attribut de l'objet du niveau 0. (voir exemple ci-dessous et page suivante).

Si A et B sont deux sous-objets de l'objet 0 et possèdent respectivement les attributs ATa et ATb alors, A et B sont par définition des sous-parties de 0 et leurs attributs sont automatiquement hérités par 0, ainsi que leurs valeurs respectives.

*Exemple* : Soit un objet PORTE doté d'un sous-objet SERRURE possédant un attribut état - pêne pouvant prendre deux valeurs : fermé, ouvert. L'objet PORTE hérite automatiquement de l'état de l'attribut du sous-objet SERRURE, il est donc raisonnable de considérer que état_pêne est également attribut de l'objet PORTE.

Ce transfert ascendant des attributs exprime les effets de l'agrégation: Tout organisme ou machine possède des organes ayant une tâche bien déterminée.

Chacun d'eux fait bénéficier l'ensemble de la machine de son travail particulier ou du simple fait d'exister (rôle structural).

La panne d'un organe se répercute souvent sur le fonctionnement global de la machine. Parfois la panne peut avoir moins d'importance aux yeux de l'utili- sateur de la machine. (fonction esthétique, suppléance par un autre organe iden- tique... )

Grâce aux partitions et aux transferts ascendant des attributs, le modèle permet de décrire le fonctionnement de toute machine avec la précision utile.

L'héritage interne conserve, si besoin est, une structure hétérogène à l'objet.

Un attribut peut intéresser une zone particulière de l'objet unique et ne pas concerner ses autres régions. Exemple : Tandis qu'une partie de l'objet reste statique une autre est mobile et transmet ce mouvement à l'objet unique.

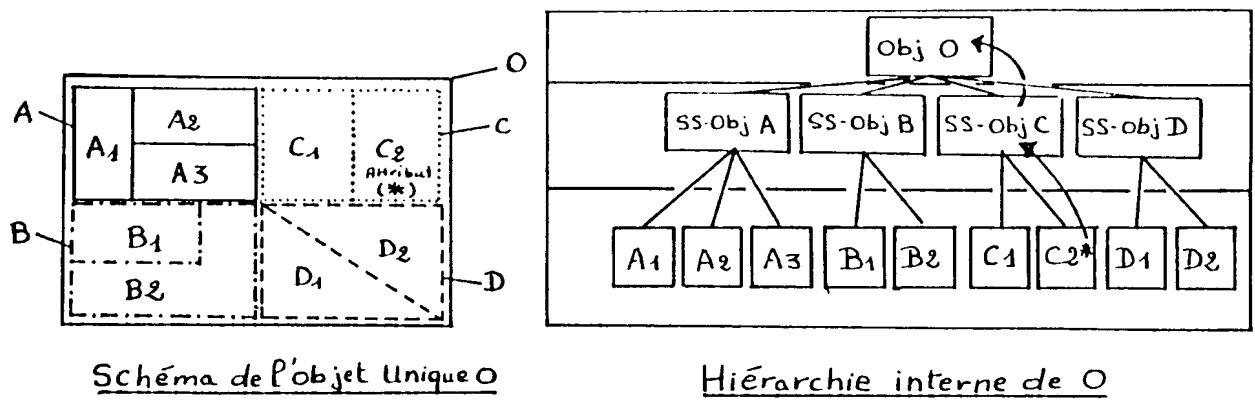

Figure 3 : Héritage interne multiple ascendant

Exemple : l'attribut de C2 (*) et sa ou ses valeurs sont hérités d'abord par le sous-objet C formé respectivement des objets C1 et C2 puis par l'objet unique 0 dont C est un des sous-objets.

Toutefois, A, B et D n'héritent pas de cet attribut.

L'héritage concerne les attributs, les fonctions internes et leurs valeurs dès qu'un sous-objet est partie de l'objet considéré.

## 2.3.5. Représentation des contraintes au niveau interne

Il s'agit soit de contraintes statiques structure, attributs, cardinalité soit de contraintes dynamiques : les fonctions traduisant l'évolution dans le temps, les fonctions définissant les actions ou méthodes réalisables par les objets. (Za-niolo, Ait-kaci et Beech 1987). Ces contraintes garantissent l'encapsulation, protègent l'objet du milieu externe en l'empêchant de prendre un état incom- patible avec son type (Moreau 1987), (Fishman, Beech et Cate 1987), (Hornick et Zdonik 1987)

*2.3.5.1. Les contraintes statiques*
1-     Les contraintes de domaine et de cardinalité
         Les contraintes de domaine fixent l'ensemble des valeurs possibles pour un attribut, une relation ou une fonction (Roche et Laurent 1986) alors que les contraintes de cardinalité établissent le nombre autorisé de valeurs simultanées.  La cardinalité s'exprime par deux chiffres : un minimum et un maximum comme dans le modèle MERISE. (Tardieu, Rochfeld et Coletti 1983).
2 -     Les contraintes d'intégrité
         Elles  définissent l 'ensemble des valeurs  que peut prendre un attribut ou une fonction *tout en restant conforme à la réalité.*

3 -     Prise en compte des contraintes statiques
         Elles sont exprimées de deux manières différentes :
         En extension  :  sous la forme d'un tableau de valeurs, quand elles sont peu nombreuses.
         *En compréhension* : en choisissant une fonction discrète ou continue définie sur un intervalle dont les deux bornes limitent l'ensemble des valeurs conservant l'intégrité.

*2.3.5.2. Les contraintes dynamiques.*

    Elles contrôlent l'évolution des objets dans le temps.
    Les modifications spontanées de leur structure, la valeur de leurs attributs doivent rester conformes à la réalité.
    Ces contraintes dynamiques sont bien traitées dans certains modèles appliqués à la bureautique privilégiant la communication. (Barbedette, Chou et Garza 1987), (Tsichritzis, Fiume et Gibbs 1987)
    Dans certains modèles, connaissance dynamique est assimilée à connaissance déductive alors que cette dernière n'a pas forcément de rapport avec le temps (voir §Expression de la connaissance déclarative), (Ferber 1986),(Vin-cent et Lermuzeaux 1984)
    Dans le modèle, les contraintes dynamiques sont exprimées à l'aide de fonctions du temps gouvernant soit la structure, soit la valeur de certains attributs.

3.3.5.2.1. EVOLUTION DE LA STRUCTURE INTERNE DES OBJETS
    Il s'agit de fonctions dynamiques, répertoriées dans la définition du type d'objet, et implantées au niveau interne des objets instances.
    Elles décrivent l'évolution de la structure d'un objet dans le temps conformément aux contraintes d'intégrité dynamiques qu'il convient d'ajuster au domaine   de définition de la fonction.

Un objet se compose de certains sous-objets à un moment, et d'autres, à un autre instant sans intervention de l'utilisateur. La structure peut varier de manière linéaire ou même périodique.

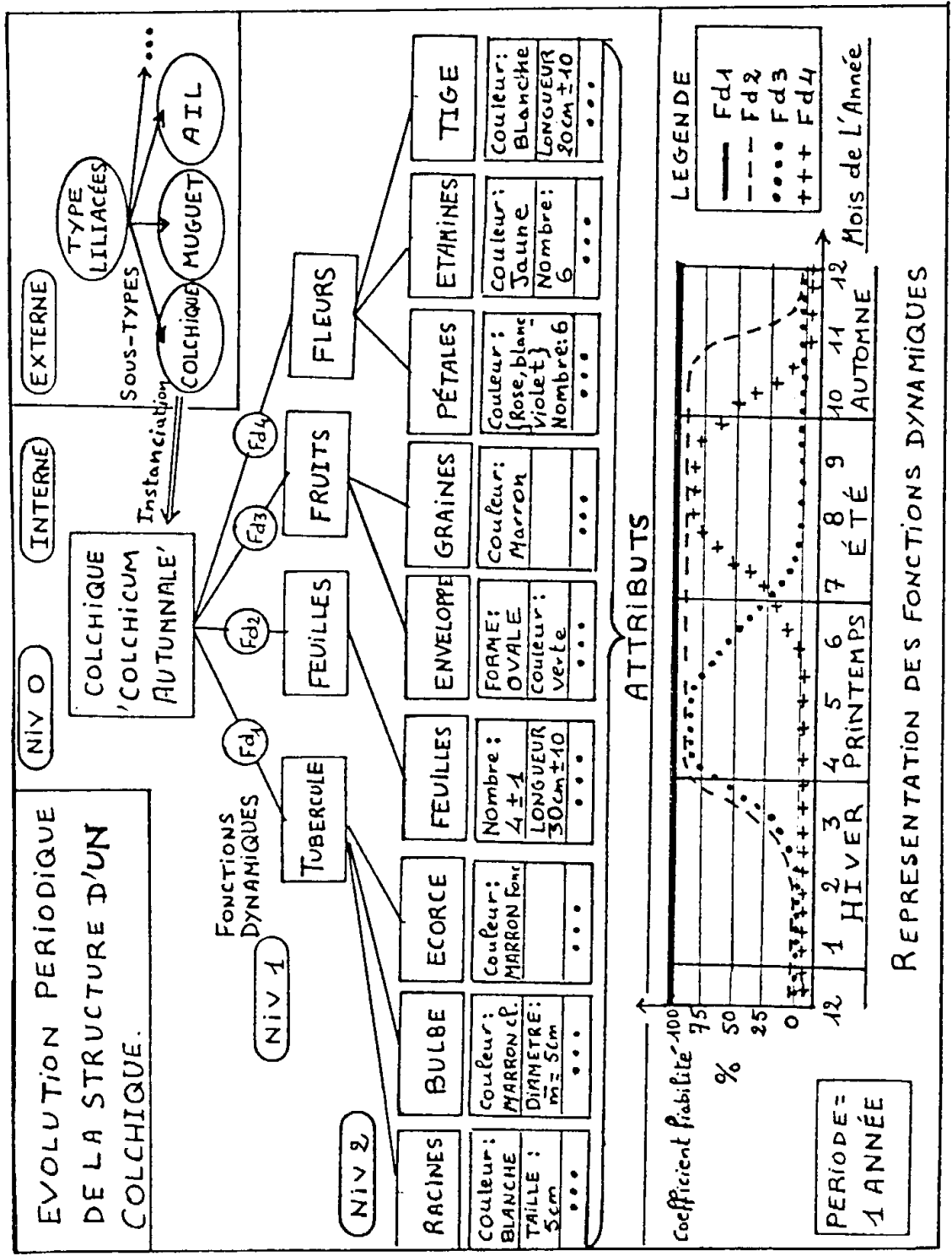

Figure 4 : Evolution de la structure interne de l'objet COLCHIQUE

*Commentaires du schéma* : Le colchique est une plante du type Liliacées, héritant, au niveau externe, des caractéristiques de cette famille de plantes. Sa structure interne change avec les saisons. La disparition de certaines parties, au cours de l'année, est traduite par les valeurs renvoyées par des fonctions dynamiques (Fdl à Fd4). Elles donnent du colchique une représentation correcte quelque soit la saison, et sans intervention de l'utilisateur.

exemple: Fd4 renvoie une valeur nulle quand le colchique est normalement dépourvu de fleur. Une valeur non-nulle, représente le coefficient de fiabilité donné par le botaniste concernant la présence de fleurs à cette époque.

Les attributs situés dans les sous-objets de niveau 2 sont transmis à l'objet unique colchique par héritage interne multiple quand les fonctions dynamiques sont non-nulles.

Le sous-objet Tubercule est toujours présent donc Fdl(t) = c = 100 %.

### 2.3.5.2.2. VARIATION DES ATTRIBUTS AVEC LE TEMPS

C'est l'évolution des valeurs des attributs et non la structure de l'objet qui est concernée. Grâce à une fonction du temps, un expert représente la variation de la valeur de référence d'un attribut au sein d'un objet exprimant une situation classique pour chaque instant.

Exemple Le taux de corticoïde dans le sang varie sur le nycthémère avec un maximum à 8h du matin elle s'abaisse dans la journée pour atteindre un minimum vers 20 h puis recommence à croître.

La valeur de référence pour un dosage dépend de l'heure de prélèvement.

### 2.3.5.2.3. EFFET DE L'EVOLUTION SUR L'HERITAGE INTERNE

Les fonctions dynamiques gouvernent la structure interne des objets.

La valeur qu'elles renvoient traduit, à tout moment, l'importance de chaque sous-objet dans l'objet qu'il contribue à former. Des fonctions linéaires, exponentielles, logarithmiques... peuvent être utilisées.

- La disparition totale d'un sous-objet s'exprime par la valeur nulle renvoyée par cette fonction.

- La croissance de cette valeur, appelée *coefficient de fiabilité* exprime la confiance accordée par l'expert à un sous-objet à un moment donné, dans la structure de l'objet unique (Colloc et Boulanger 1987). La fonction dynamique de structure et sa valeur sont immédiatement répercutées vers les objets pères, jusqu'à l'objet unique.

Le coefficient de fiabilité intervient souvent comme paramètre des fonctions d'évaluation spécifiques d'un sous-objet donné.

Les fonctions de structure et les attributs fonction du temps sont recalculées périodiquement et leurs valeurs sont héritées automatiquement (figure 4).

*2.3.6. Expression de la connaissance déclarative.*

Les connaissances déclaratives représentent les faits grâce aux attributs et à la structure d'un objet unique : nombre de niveaux et la nature des sous-objets intervenant dans sa constitution.

### 2.3.6.1. Les attributs

Les attributs sont implantés au niveau interne et donc protégés du milieu externe. Leurs valeurs sont souvent exploitées par des fonctions internes. Ce ne sont pas des sous-objets mais bien des champs de données mono ou multi-valués chargés de les décrire. (Penedo 1986), (Gallo, Minot et Thomas 1986), (King 1986), (Koskimies et Paakki 1987).

Les attributs sont atomiques et leur structure est plane, ce qui les différencie des objets feuilles, c'est qu'ils ne peuvent avoir d'attribut alors que les sous-objets feuilles en ont au moins un.

Les attributs sont situés nécessairement dans les sous-objets feuilles au niveau le plus bas de l'arborescence interne.

Ils peuvent être représentés par une fonction discrète ou continue, avec un ou plusieurs paramètres dont éventuellement le temps.

### 2.3.6.2. Les attributs qualifiants : leur rôle interne

Il s'agit d'attributs particuliers répertoriés au niveau externe comme caractéristiques du type d'objet et dont l'instanciation est obligatoire au niveau interne.

Tous les attributs d'un objet ne sont pas qualifiants, l'attribution d'une valeur est alors facultative. La contrainte de cardinalité des attributs qualifiants est telle que $min > 0$ et $max >= min$.

### Les fonctions

Les fonctions, faisant partie intégrante des objets, définissent un attachement procédural, celles contribuant à leur gestion appartiennent au niveau externe : fonctions d'évaluation globales et macrofonctions du métalangage. (voir conclusion).

Le modèle propose l'utilisation de fonctions internes continues et discrètes :

- fonctions dynamiques de structure, fonctions exprimant l'évolution d'un attribut, les fonctions d'évaluation propres à l'objet, des fonctions chargées de tâches particulières (calcul, dessins...).

Ces fonctions internes sont comparables à des méthodes activées par un message adapté reçu par l'objet. (Roche et Laurent 1986), (Methfessel 1987), (Fishman, Beech et Cate 1987).

Dotées d'un ou plusieurs paramètres, elles renvoient éventuellement un message destiné au niveau externe.

Protégées du milieu externe, ces fonctions revêtent des formes variées selon les applications. (Voir exemple § Les fonctions d'évaluation propres à l'objet)

### 2.3.7. Représentation de la connaissance déclarative

Par opposition à la connaissance déclarative, la *connaissance déductive* provient d'une action ou d'une opération (d'un calcul) sur des faits.

Dans le modèle, elle est représentée à l'aide de fonctions d'évaluation comparant l'état des objets. Il en existe de deux sortes :

*2.3.7.1. Les fonctions d'évaluation propres à l'objet*

Internes à un objet, elles renvoient au niveau externe à la fois une image de la structure et de l'état de l'objet.

Un système expert peut exploiter cette image en la comparant à celle d'autres instances du même type, (de structure analogue, et d'état différent), à l'aide des fonctions d'évaluation globales (§ Les fonctions d'évaluation globales). Les fonctions d'évaluation internes (f.e.i.) traitent d'une manière très souple l'incertitude des seuils théoriques en s'affranchissant de la dichotomie (valeur d'attribut inférieure ou supérieure au seuil).

La f.e.i. a comme paramètre en général les attributs d'un sous-objet et compare des valeurs de référence, fournies par l'expert (objet théorique) avec celles constatées (objet réel), elle renvoie une valeur d'autant plus faible qu'il y'aura identité entre les attributs des sous-objets comparés. On peut construire plusieurs f.e.i. sur les mêmes attributs selon le but visé par l'expert.

Il s'agit le plus souvent de fonctions continues, mais parfois discrètes quand le nombre d'états distincts possibles pour les attributs est fini. (Colloc et Boulanger 1987)

Exemple: la fonction d'évaluation interne ci-dessous permet de comparer la valeur de l'attribut fièvre dans deux objets : l'un représentant la maladie du patient, l'autre une maladie de la base de connaissances.

D (en %) = ( | V malade - V théorique | * Facteur échelle)

D Fièvre (en %) (|Température malade – 39° | * 20)

F. Echelle = 100 / Max ( (V théor - Min signe),(V théor - Max signe) ).

Plus les attributs sont proches, plus la valeur retournée est faible. (0% : parfaitement identique, 100% : valeur extrême).

Cette valeur peut intervenir comme paramètre d'une fonction d'évaluation globale.

*2.3.7.2. Les fonctions d'évaluation globales*

Attachées à un type, donc au niveau externe (voir § Le niveau externe), elles évaluent toutes les instances d'un type d'objet de la base de connaissances.

Les résultats obtenus sont fournis à l'utilisateur, et exploités, au niveau externe, pour effectuer des diagnostics ou prendre des décisions. Ces fonctions sont le plus souvent discrètes.

2.4. Le niveau externe

Les classifications regroupent dans le même ensemble des objets se ressemblant plus où moins. On ne devrait employer ce terme qu'au pluriel car on peut en bâtir une infinité.

Il s'agit de regrouper des objets ayant un ou plusieurs points communs dans des familles nommées classes comme dans le langage SMALLTALK-80. (Goldberg et Robson 1983), (Mevel et Guegen 1987).

Ces classes peuvent elles mêmes être subdivisées en sous-classes d'objets selon les ressemblances qui peuvent encore être remarquées entre les objets de la classe. A un moment donné, les objets n'ayant plus de point commun, il devient impossible de créer des sous-classes regroupant plus d'un objet. C'est à ce point que la hiérarchie de classes rencontre la hiérarchie interne de l'objet unique.

Le niveau externe est gouverné par les concepts de généralisation et de spécialisation.

## 2.4.1. Généralisation et Hiérarchie de types et de classes : hiérarchie externe

On distingue deux types de généralisation : la généralisation d'objets atomiques ou généralisation d'instance ou classification et la généralisation d'objets moléculaires ou de classes s'appliquant récursivement à plusieurs niveaux (Codd 1979).

La généralisation bâtit une hiérarchie de type à l'aide de l'association EST-UN, la spécialisation est son inverse (Smith et Smith 1977). Pour chaque type d'objet, il existe une classe regroupant les objets instances répondant aux exi- gences du type.

C'est seulement quand la spécialisation est *totale* qu'elle établit une partition sur la classe, dont les parties sont appelées sous-classes. (Bouzeghoub 1987)

Soit TyA et TyB deux types, si on a TyB --EST-UN--> TyA signifie que les objets de type TyB sont aussi de type TyA, on dit que TyB est un « sous-type » de TyA et que TyA est le « super-type » de TyB.

Quand un type ne peut avoir qu'un seul super-type, il dépend d'une hiérarchie externe simple de représentation arborescente (Perrot 1987). A l'inverse de certains travaux (Banerjee, Chou et Garza 1987), (Zdonik et Wegner 1985), nous nous situons dans le cas où un type ne peut avoir plusieurs super-types et par conséquent il n'existe pas dans notre modèle *de hiérarchie externe multiple*.

## 2.4.2. Définition d'un type d'objets

Il recense les caractéristiques statiques, dynamiques et procédurales qui doivent être implantées au niveau interne des objets uniques pour être conforme au type. Ces caractéristiques sont héritées du niveau externe vers le niveau interne (instanciation).

On peut définir un type d'objets de référence TY0 comme suit

1 -    *Nom du type d'objets*

2 -    *Les caractéristiques statiques*

* Des propriétés structurales : le nombre de niveaux Nbnv de l'arborescence de sous-objets nécessaire à l'expert pour une représentation correcte. (NB: dans l'absolu Nbnv $\rightarrow \infty$ et plus Nbnv est grand, meilleure est la précision de défini- tion des objets).

L'expert choisira Nbnv selon la précision désirée.

* La liste des attributs qualifiants dont l'instanciation est obligatoire pour chaque objet du type. C'est l'ensemble des attributs qualifiants transmis par héritage externe aux sous-types et à leurs objets instances.

3 -    *Les caractéristiques dynamiques*

Les fonctions dynamiques de structure f(t). (voir § Les contraintes dyna-miques). La prépondérance de certains liens sur d'autres est exprimée par des fonctions internes.

\* Les attributs qualifiants exprimés par une fonction du temps.

4 - *Les procédures*

\*        Fonctions d'évaluation globales liées au type donc du niveau externe.

\*        Les fonctions réalisant des tâches spécifiques activées à la réception d'un message provenant d'un autre objet ou d'une interface avec l'utilisateur.

5 - Les types des sous-objets

De même que le type des objets de niveau 0, on définit récursivement les types des sous-objets dont le niveau est compris entre 1 et 9.

N.B. Quelque soit niv > 0 : TYniv n'est pas un sous-type de TY0.

*Exemple* : Une voiture Vl appartient à la Classe VOITURES (EXTERNE). Elle a comme sous-objet de niveau niv (/niv > 0) une portière pi qui fait partie de l'objet unique Vl (INTERNE) et appartient à la Classe des PORTIERES elle même sous-classe de la Classe des PORTES.(EXTERNE) Il est évident que l'objet portière n'est pas instance de la Classe VOITURES car ce n'est pas une voiture.(Elle n'en a ni la structure ni les attributs qualifiants).

Par contre la portière est un sous-objet de l'objet voiture. La portière est du type PORTES.(EXTERNE) Toutefois le type d'objets PORTIERES est plus précis et impose une structure et des attributs qualifiants suplémentaires par rapport au type d'objet PORTES. (ex: Une portière a une vitre coulissante ce qui est inhabituel pour une Porte.)

L'exemple de la voiture est traité par de nombreux auteurs : (Hornick et Zdonik 1987), (Banerjee, Chou et Garza 1987), (Ogonowski 1984).

### 2.4.3. La classe d'objets

C'est l'ensemble des objets instances d'un type d'objets considéré. Les objets appartenant à la classe bénéficient des propriétés structurales et des contraintes d'intégrité définies dans le type d'objets associé.

### 2.4.4. Les sous-classes d'objets

Ensemble des objets d'un sous-type considéré.

Soit Cl une classe d'objets et C2 une autre classe d'objets on dit que C2 est une sous-classe d'objets de Cl si et seulement si C2 est inclus dans Cl c'est-à-dire : quelque soit obj : obj $\in$ C2 $\Rightarrow$ obj $\in$ Ci.

### 2.4.5. Héritage externe simple

A la notion d'hiérarchie simple se superpose celle d'héritage simple. C'est la faculté pour un type de bénéficier des caractéristiques statiques : structure, propriétés, attributs mais aussi dynamiques méthodes, fonctions, opérations... provenant d'un supertype unique. Ce transfert a lieu de haut en bas suivant la hiérarchie de types.

*2.4.6. Instanciation des types d'objets existant.*

La figure ci-dessous, reprend l'exemple du stylo à bille de la page 5. Elle montre le niveau externe où résident les types organisés hiérarchiquement, l'arborescence interne des sous-objets du stylo à bille et l'instanciation des types.

N.B. Au niveau externe, les types des sous-objets feuilles (niveau 2) n'ont pas été représentées sur le schéma pour qu'il reste lisible.

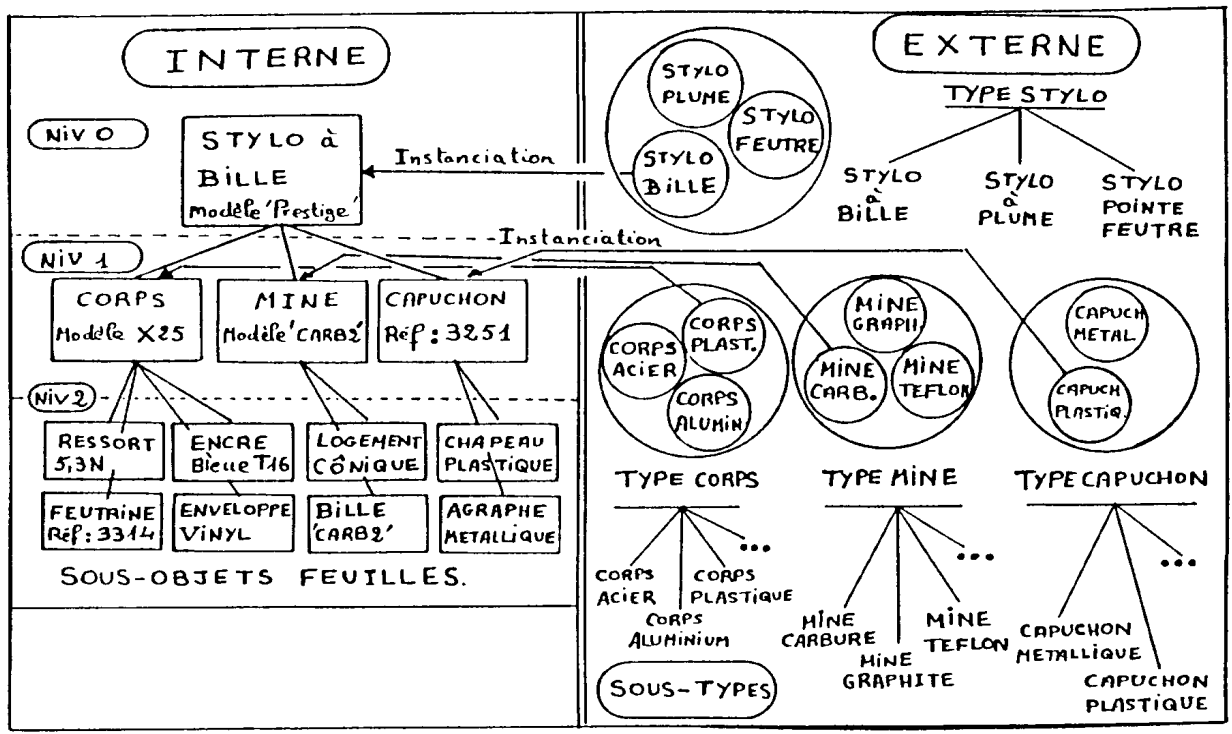

Figure 5 : Présentation des deux niveaux conceptuels et instanciation

## 3. Conclusion et recherches futures

Nous avons présenté un modèle définissant les concepts nécessaires à la re-présentation de connaissances empiriques sous forme d'objets.

Les points forts de ce modèle étant la distinction, entre le niveau interne et le niveau externe de représentation, l'héritage interne ascendant, la notion d'at-tribut qualifiant et les fonctions d'évaluation ainsi que les fonctions dynamiques.

Actuellement nous établissons une méthode pour construire un système expert exploitant les connaissances représentées à l'aide de ce modèle. Nous construisons également des outils permettant : la gestion d'une base d'objets (création, suppression, mise à jour, versions...), l'interface avec l'utilisateur, la réalisation de systèmes experts.

Nous essayons, à présent, de définir des "primitives" de la méthode permet-tant la mise en oeuvre des concepts du modèle à objet, en cours de réalisation : Messages, Métalangage, Métafonctions et Versions.

* Les messages sont définis comme un moyen de communication entre les objets. Ils ne sont pas compréhensibles par tous les objets mais uniquement ceux auxquels ils sont destinés. (Narat et Lochet 1987). Un message déclenche une fonction au niveau interne des objets destinataires.

La notion de message dans les modèles objets est analogue au fonctionnement des hormones dans l'organisme humain.

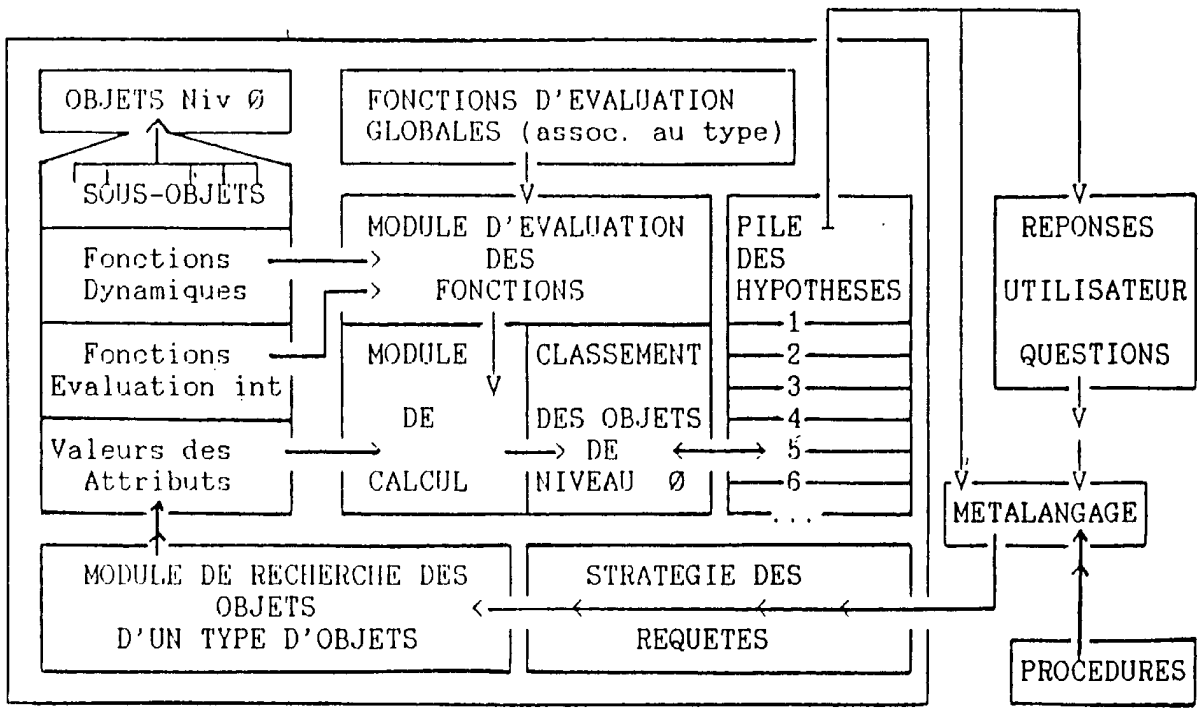

Figure 6 : Schéma décrivant le fonctionnement du moteur de raisonnement

* Le métalangage permet de gérer différents types d'objet nécessaires à la résolution d'un problème donné.

Il se comporte comme un interpréteur des commandes exprimées au niveau EXTERNE.

Il expédie des messages déclenchant l'exécution de méthodes au sein d'objets cibles d'un type donné.

Il peut agir globalement sur toutes les instances d'un type ou sur des objets individuellement.

Le niveau méta stocke les messages réponses des objets.

Il permet de bâtir des "métafonctions" permettant de réaliser des tâches complexes comme Le diagnostic, pronostic, simulation, comparaison de stratégies...

Certaines de ces métafonctions exploitent des algorithmes permettant de rechercher des objets dans l'arborescence interne.

* Pour nous, la gestion des versions d'un objet a toujours lieu au niveau interne et l'on distingue deux catégories de modifications possibles :

1) *Les modifications spontanées*

- Les fonctions dynamiques implantées au niveau interne décrivant l'évolution de la'structure de l'objet.

- Les attributs fonctions du temps exprime l'évolution des valeurs de références.

Il est donc facile de calculer la structure et l'état des attributs d'un objet à un moment donné.

On peut également prévoir les structures et les états futurs de l'objet. Cette faculté originale est précieuse en simulation.

Il s'agit ici d'une véritable cinématique de l'objet et non pas simplement de la sauvegarde de clichés successifs. (Adiba 1981)

2) *Les modifications d'un objet par l'utilisateur*

Elles doivent être compatibles avec le type de l'objet et partant respec- ter les contraintes d'intégrité et de cardinalité afin de conserver une valeur légale aux attributs.

## *** Bibliographie ***

Adiba M., « Derived Relations : A Unified Mechanism for Views, Snapshots and Distributed Data », *VLDB*, Cannes, France, 1981.

Banerjee J. , Chou hong-tai, Garza Jorge. F., « Data Model Issues for Object-Oriented Applications. », ACM *Transactions on Office Information Systems*, Vol 5, N°l, January 1987, pages 3-26.

Barbedette G., Richard P., « Tool: A Typed Object-Oriented Lanquage for Data Bases. », PRC-BD3 Port Camargue ed. I.N.R.I.A. *Journées Bases de Données Avancées*, mai 1987.

Boulanger D., Morejon J., « Extension d'un Modèle Entité-Relation par les concepts de Généralisation et d'Héritage. », *Journées Internationales des Systèmes Informatiques, JISI'88* , Tunis, avril 1988.

Bouzeghoub M., « Les contraintes d'héritage dans les hiérarchies de généralisation. », *Journées FIRTECH Systèmes et Télématique Bases de Données et Intelligence artificielle,* Paris, 9-10 Avril 87, pp 19-35.

Brodie M.L., « On Modellinq Behavioural Semantics of Databases », *VLDB*, Cannes, France, 1981.

Codd E.F., « Extending the Data Base Relational Model to Capture more Mea- ning », *ACM transaction on data base systems*, Vol. 4, n'4, 1979.

Colloc J., Boulanger D., « Conception d'un Système Expert Orienté Objet des- tiné à l'aide au Diagnostic Médical. », *Colloque Intelligence Artificielle et Santé SITEF, Toulouse*, 2-3 Octobre 1987.

Dittrich K.R., Lorie R.A., « Object-oriented database concepts for engineering applications. », CH2136-0/85/0000/0321 01.00 1985 *IEEE COMPINT 85.*

Ferber J., « Systèmes experts et approches orientées objets. », Sixièmes Journées Internationales d'Avignon pp. 525-542.

Fishman D.H., Beech D., Cate H.P., « IRIS: An Object-Oriented Database Management System. », *ACM Transactions on Office Information Systems*, Vol 5, N°l, January 1987, pp. 48-69.

Flory A., Bertin M. « Un outil pour le contrôle sémantique de la cohérence des systèmes de connaissances », *Journées Bases de Données de 3ème Génération, Giens*, Mai 1986.

Gallo F, Minot R., Thomas I., « The object Management System of PCTE as a Software Engineering Database Management System. », *Software Engineering Symposium on Practical Software Development Environments*, ed. by, P. Henderson, déc. 1986.

Goldberg A., Robson D., Smalltalk-80 : The Lanquage and its Implementation, Addison-Wesley, Reading, Mass., 1983.

Hornick M.F., Zdonik S.B., "A Shared Segmented Memory System for an Object-Oriented Database." *ACM Transactions on Office Information Systems*, Vol. 5, N°l, January 1987, Pages 70-95.

King R., *A Database Management System based on an Object-Oriented Model,* Expert Database Systems, Larry Kerschberg Editor, Benjamin/ Cummings Publishing Company Inc, 1986.

Koskimies K., Paakki J., « TOOLS: A Unifying Approach to Object-Oriented Language Interpretation. », *ACM* 0-89791-235-7/87/0006/0153

Laurent P., « Systèmes Experts et Langages Orientés Objets: Un mariage fruc-tueux. », *MICRO-SYSTEMES*, juillet-Août 87, pp 145-149.

Methfessel R., « Implementing an access and object oriented paradigm in a lan-guage that supports neither. », *SIGPLAN NOTICES V22* / 4, April 1987.

Mevel A., Guegen T., « SMALLTALK-80 », Ed. Eyrolles, 61 Bd St Germain 75005 Paris, 1987.

Moreau R., « Architecture des machines et langages de programmation », *Journée "Les langages de programmation et leur évolution"*, groupe ESIEE, juin 1987.

Narat V., Lochet P.Y.,« Les différentes techniques de représentation de connais-sances utilisées en intelligence artificielle.» *Modèle et Bases de Données,* n°6, juin 1987.pp 25-36.

Ogonowski A., « Les langages de programmation Orientés Objets.», *Colloque* d'Intelligence Artificielle d'Aix en Provence, 17-21 Septembre 1984, publication du GR-22.

Penedo M.H., « Prototyping a Project Master Data Base for Software Engineering Environments. » *ACM SIGSOFT/SIGPLAN*, Software Engineering Sympo- sium on Practical Software Development Environments, dec. 1986.

Perrot J.F., "Les langages à objets", Journée "*Les langages de programmation et leur évolution*", *groupe ESIEE*, juin 1987.

Roche C., Laurent J.P., « LRO 2 – Artificial Intelligence and Object Oriented Lanquages », *Information Processing*, 86 H-J. Kugler (Ed.) Elsevier Science Publishers BV (North Holland), Ifip 1986.

Smith J.M., Smith D.C.P., « Database Abstractions : Aggregation and Generalization. » , *ACM Transactions on Office Information Systems*, Vol 2, n°2, june 1987, pp. 105-133, 1987

Tardieu H., Rochfeld A., Coletti R. , « *La Méthode MERISE Principes et Outils* », Les Editions d'Organisation Paris, 1983.

Tsichritzis D., Fiume E., Gibbs S., « *KNOs Knowledge Acquisition Dissemination and Manipulation Objects* » *ACM Transactions on Office Information Systems*, Vol 5, N'l, January 1987, Pages 96-112.

Vincent D., Lermuzeaux J.M., « ROSACE:UN système de gestion de base de connaissances orienté objet et ANAEL-SESAME : Son application à la communication homme-machine avec un système bureautique. *», Colloque d'Intel- ligence Artificielle Aix-en-Provence*, 17-21 Septembre 1984.

Zaniolo C., Ait-kaci H., Beech D. & cie , "Object Oriented Database Systems and Knowledge Systems" EXPERT DATABASE SYSTEMS P50-P65.

Zdonik S.B., Wegner P., "Language and Methodology for Object-Oriented Database Environments.", BROWN UNIVERSITY, Dept of Computer Science, TECHNICAL REPORT N' CS-85-19 NOVEMBER 1985.